\documentclass{article}

\usepackage[preprint]{neurips_2020}
\usepackage{amsmath}
\usepackage[pdftex]{graphicx}

\usepackage[utf8]{inputenc} 
\usepackage[T1]{fontenc}    
\usepackage{hyperref}       
\usepackage{url}            
\usepackage{booktabs}       
\usepackage{amsfonts}       
\usepackage{nicefrac}       
\usepackage{microtype}      

\title{The Golden Ratio of Learning and Momentum}

\author{Stefan Jaeger\\
National Library of Medicine\\
National Institutes of Health\\
Bethesda, MD 20894, USA\\
\texttt{stefan.jaeger@nih.gov}
}

\begin{document}

\maketitle

\begin{abstract}
Gradient descent has been a central training principle for artificial neural networks from the early beginnings to today's deep learning networks. The most common implementation is the backpropagation algorithm for training feed-forward neural networks in a supervised fashion. Backpropagation involves computing the gradient of a loss function, with respect to the weights of the network, to update the weights and thus minimize loss. Although the mean square error is often used as a loss function, the general stochastic gradient descent principle does not immediately connect with a specific loss function. Another drawback of backpropagation has been the search for optimal values of two important training parameters, learning rate and momentum weight, which are determined empirically in most systems. The learning rate specifies the step size towards a minimum of the loss function when following the gradient, while the momentum weight considers previous weight changes when updating current weights. Using both parameters in conjunction with each other is generally accepted as a means to improving training, although their specific values do not follow immediately from standard backpropagation theory. This paper proposes a new information-theoretical loss function motivated by neural signal processing in a synapse. The new loss function implies a specific learning rate and momentum weight, leading to empirical parameters often used in practice. The proposed framework also provides a more formal explanation of the momentum term and its smoothing effect on the training process. All results taken together show that loss, learning rate, and momentum are closely connected. To support these theoretical findings, experiments for handwritten digit recognition show the practical usefulness of the proposed loss function and training parameters.
\end{abstract}


\section{Introduction}
\label{sec:Introduction}
Artificial neural networks (ANNs) have been at the center of machine learning and artificial intelligence from the early beginning.
However, the basic training principle of ANNs has not changed since the inception of feed-forward, multilayer networks~\citep{rumelhart1986learning}. While their structure and depth have been constantly developed further, leading to the modern deep learning networks, training of parameters is traditionally based on gradient descent and backpropagation. This paper revisits this common way of training ANNs with backpropagation, arguing that it should rather be considered as a two-way relationship between input and output rather than a one-sided teaching process. Following this train of thought, and motivated by biological learning systems, this paper develops a computational model for which learning is tantamount to making an observed input identical to the actual input, implying a difference between perception and reality. This boils down to finding the optimal gradient of an information-theoretical loss function, where the search space includes points for which the synaptic input coincides with the output. The theory developed in this paper will show that the golden ratio plays a central role in this learning process, defining points of minimum uncertainty, which define reality. Furthermore, the golden ratio allows to derive theoretical weights for the learning rate and momentum weight. The paper shows that these values match closely the values traditionally used in the literature, which are determined empirically. To provide further evidence that the presented theoretical framework is of practical significance, a practical test is carried out in the last section of this paper. For this test, a deep learning network is applied to handwritten digit recognition, using the proposed loss function and learning parameters.

The paper is structured as follows:
Section~\ref{sec:Backpropagation} briefly summarizes the basic principle of backpropagation, including the momentum term. Section~\ref{sec:SynapticTransmissionProcess} highlights the basic mechanism of natural synaptic signal processing based on which Section~\ref{sec:ComputationalModel} then derives a computational learning model. In Section~\ref{sec:GoldenRatio}, the golden ratio and its mathematical definition are highlighted. Section~\ref{sec:LossFunction} presents the information-theoretical loss function and develops the regularization of momentum. Finally, Section~\ref{sec:ExperimentalEvaluation} shows experimental results before a conclusion summarizes the paper.

\section{Backpropagation}
\label{sec:Backpropagation}
In more than thirty years, backpropagation has established itself as the commonly used optimization principle for supervised learning with ANNs, including modern deep learning networks~\citep{rumelhart1986learning,bengio2012practical}. Backpropagation adjust the weights in ANNs so that the difference between network output and teaching input becomes smaller.

\subsection{Basic principle}
\label{subsec:BasicPrinciple}
The backpropagation algorithm is a gradient descent method that starts by computing the gradient of the loss function defining the difference between the network output and the teaching input~\citep{lecun2012efficient}. A commonly used loss function~$L$ is the sum of the squared error (SSE) between the network predictions~$Y$ and training targets~$T$~\citep{widrow2019nature}:
\begin{equation}
L = \frac{1}{N} \sum_{n}^{N} \sum_{k}^{K} \ (Y_{nk}-T_{nk})^2,
\label{sseLossEquation}
\end{equation}
where~$N$ is the number of observations and~$K$ is the number of classes. The gradient is computed with respect to each network weight. It can be computed one layer at a time, using the chain rule, starting at the output layer and then iterating backwards. Propagating the gradient back through the network for each weight is what gives this method its name. An important term in this backpropagation process is the derivative of the loss function with respect to the predictions~$Y$. For the SSE loss function, this derivative takes the following form for an observation vector~$Y$ and a training vector~$T$:
\begin{equation}
\frac{dL}{dY} = \frac{2 \cdot (Y-T)}{N}
\label{backwardLossEquation}
\end{equation}
The last step in the backpropagation process is to move along the gradient towards the optimum by adjusting each weight $w_{ij}$ between two nodes, $i$ and $j$, in the network. This is achieved by adding a $\Delta w_{ij}$ to each weight $w_{ij}$, which is the partial derivative of the loss function~$L$ multiplied by the so-called learning rate~$\eta$ (multiplied by $-1$ to move towards the minimum):
\begin{equation}
\Delta w_{ij} = -\eta \frac{\partial L}{\partial w_{ij}}
\label{deltaEquation}
\end{equation}
The learning rate~$\eta$ is a parameter that had to be determined largely empirically so far~\citep{bengio2012practical}. In setting a learning rate, there is a trade-off between the rate of convergence and the risk of passing over the optimal value. A typical value for the learning rate used in practice appears to be~$0.01$, although reported values have ranged across several orders of magnitude.

\subsection{Momentum term}
\label{subsec:MomentumTerm}
Updating weights by adding the delta of Eq.~\ref{deltaEquation} does not guarantee reaching the global optimum. In practical experiments, adding a so-called momentum term has proved to be effective in improving performance. The momentum term for a weight $w_{ij}$ in the network is its delta from the previous iteration, $t-1$, multiplied with a weighting factor~$\alpha$. With momentum, the delta term in Eq.~\ref{deltaEquation} becomes
\begin{equation}
\Delta w_{ij}(t) = -\eta \frac{\partial L}{\partial w_{ij}(t)} + \alpha \cdot \Delta w_{ij}(t-1)
\label{deltaWithMomentumEquation}
\end{equation}
The common understanding is that the momentum term helps in accelerating stochastic gradient descent (SGD) by dampening oscillations. However, it introduces another parameter~$\alpha$ for which only empirical guidelines exist but no theoretical derivation of its value, although second order methods have been tried~\citep{bengio2012practical,sutskever2013importance}.

Similar to the learning rate~$\eta$, several values have been tried out in practice for the momentum weight~$\alpha$, although values around $0.9$ seem to be more commonly used than others.

\section{Synaptic Transmission Process}
\label{sec:SynapticTransmissionProcess}
The basic building blocks of the human brain are neurons, which are connected and communicate with each other via synapses. It is estimated that the human brain has around $8.6 \times 10^{10}$ neurons~\citep{herculano2009human}, which can each have several thousand synaptic connections to other neurons, and that the number of synapses ranges from $10^{14}$ to $5 \times 10^{14}$ for an adult brain~\citep{drachman2005we}. A neuron sends a signal to another neuron through its axon, which is a protrusion with potentially thousands of synapses and which can extend to other neurons in distant parts of the body. The other neuron receives the signal via so-called dendrites that conduct the received signal to the neuron's body.

\begin{figure}[ht] 
\begin{center}
\centerline{\includegraphics[width=0.7\columnwidth]{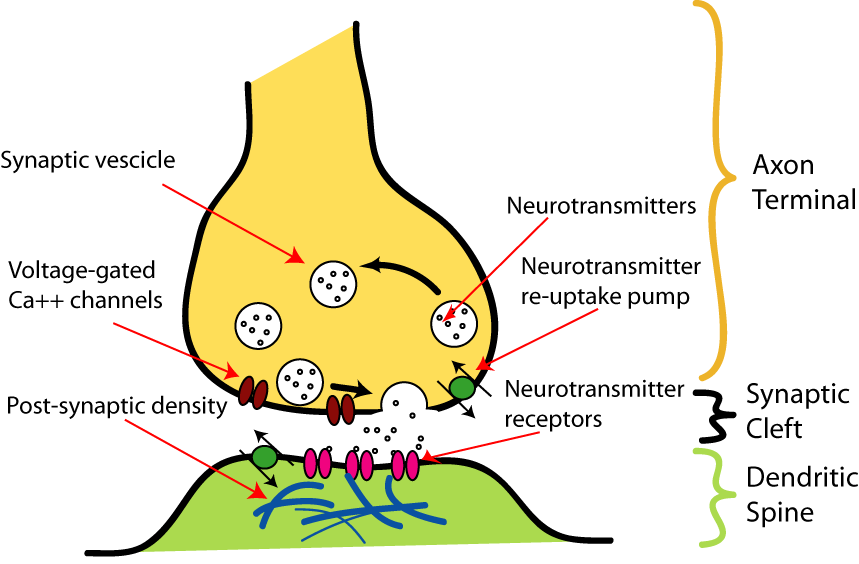}}
\caption{Signal transmission at a chemical synapse~\citep{julien2005} (Source: Wikipedia)}
\label{synapseFigure}
\end{center}
\end{figure}

A synapse is a membrane-to-membrane interface between two neurons that allows either chemical or electrical signal transmission~\citep{lodish2000neurotransmitters}. In case of a chemical synapse, the signal is transmitted by molecular means from the presynaptic axon terminal of the sending neuron to the postsynaptic dentritic terminal of the receiving neuron. This is accomplished by neurotransmitters, which can bridge the synaptic cleft, a small gap between the membranes of two neurons, as illustrated in Figure~\ref{synapseFigure}.
The small volume of the synaptic cleft allows neurotransmitter concentration to increase and decrease rapidly.

Although synaptic processes are not fully understood, it is believed that they are at the heart of learning and memorizing patterns. It is known that the signal transmission at a chemical synapse happens in several steps. Except for the last step, each step takes no more than a fraction of a millisecond. The transmission is first triggered by an electrochemical excitation (action potential) at the presynaptic terminal. This excitation then causes calcium channels to open, allowing calcium ions (Ca++) to flow into the presynaptic terminal. The increased concentration of calcium ions in the presynaptic terminal causes the release of neurotransmitters into the synaptic cleft. Some of these neurotransmitters then bind to receptors of the postsynaptic terminal, which opens ion channels in the postsynaptic membrane, allowing ions to flow into or out of the postsynaptic neuron. This changes the transmembrane potential, leading to an excitation or inhibition of the postsynaptic neuron. Eventually, the docked neurotransmitters will break away from the postsynaptic receptors. Some of them will be reabsorbed by the presynaptic neuron to initiate another transmission cycle. 

The computational model developed in the next section is based on the assumption that the concentration of calcium ions~$p$ largely defines the strength of the signal transmitted. Furthermore, the assumption is that the strength of the signal is defined by the relation between the concentrations of calcium ions inside the presynaptic terminal ($1-p$) and outside the terminal ($p$). For example, when all calcium ion channels are closed, the outside concentration will be $p=1$ and there will be no signal transmitted: $(1-p)/p = 0$. On the other hand, when the ion channels open up because of an excitation of the presynaptic terminal, calcium ions rush into the terminal due to a greater extracellular ion concentration. The maximum signal strength of~$1$ will be reached for $p=0.5$, when the concentrations of calcium ions inside and outside the terminal are in equilibrium.

\section{Computational Model}
\label{sec:ComputationalModel}
The computational model set forth here is based on the expected information that is either received or generated by the presynaptic neuron~\citep{hodgkin1990quantitative}. This section proceeds from the common conception that information is a measure for uncertainty, and is related to energy, without going into details about how information relates to energy in physics. The terms {\it information} and {\it energy} are used synonymously in the following. In a traditional information-theoretical approach, the informational content of a probabilistic event is measured by the negative logarithm of its probability, which is then used to compute the expected information or {\it entropy}~\citep{shannon1948mathematical}. This section will follow along these lines.

Building on Section~\ref{sec:SynapticTransmissionProcess}, the information conveyed by the signal $(1-p)/p$ is computed as ${-\ln((1-p)/p)}$. The expected information or energy for this signal is then computed as the product of its information and "probability," where the latter can be viewed as the likelihood of a calcium ion entering the presynaptic terminal through an ion channel: 
\begin{equation}
E = -p \cdot \ln\bigg(\frac{1-p}{p}\bigg),
\label{computationalModelEquation}
\end{equation}
with $p \in \big[\frac{1}{2}\,;1\big]$~\citep{jaeger2013neurological}.

Developing the model further, the signal $(1-p)/p$ can be regarded as the signal perceived by the presynaptic neuron, whereas~$p$ is the actual signal coming from outside the neuron. Continuing this thought leads to the observation that reality is defined by the agreement between the perceived signal and the actual signal. Both signals are identical when $(1-p)/p$ equals~$p$, which is the case when $p$ is equal to the golden ratio~\citep{livioGoldenRatioBook,weiss2003golden}. Therefore, the golden ratio defines points for which the perceived and the actual signal coincide. Section~\ref{sec:GoldenRatio} will discuss the golden ratio in more detail, as it is central to this paper.

Assuming that the perceived signal $(1-p)/p$ is equal to the actual signal~$p$, the formula for the expected information, as given by Eq.~\ref{computationalModelEquation}, can be transformed into a symmetric equation as follows:
\begin{eqnarray}
\nonumber E & = & -p \cdot \ln\bigg(\frac{1-p}{p}\bigg) \\
& \Leftrightarrow & -p \cdot \ln\Big(1-p^2\Big) \\
& \Leftrightarrow & -p \cdot \ln\Big(\sqrt{1-p^2}\Big) \cdot 2 \\
& \Leftrightarrow & -\sin(\phi) \cdot \ln\big(\cos(\phi)\big) \cdot 2,
\label{pythagoreanEquation}
\end{eqnarray}
where the last expression holds for an angle~$\phi \in \big[0\,;\frac{\pi}{2}\big]$. Note that this last expression is symmetric in that we can swap sine and cosine to obtain the expected information for a signal sent in the opposite direction. The signal in the forward direction corresponds to the traditional forward pass in a feed-forward neural network, whereas the signal in the opposite direction represents the signal generated by the teaching input.

\begin{figure}[ht] 
\begin{center}
\centerline{\includegraphics[width=0.7\columnwidth]{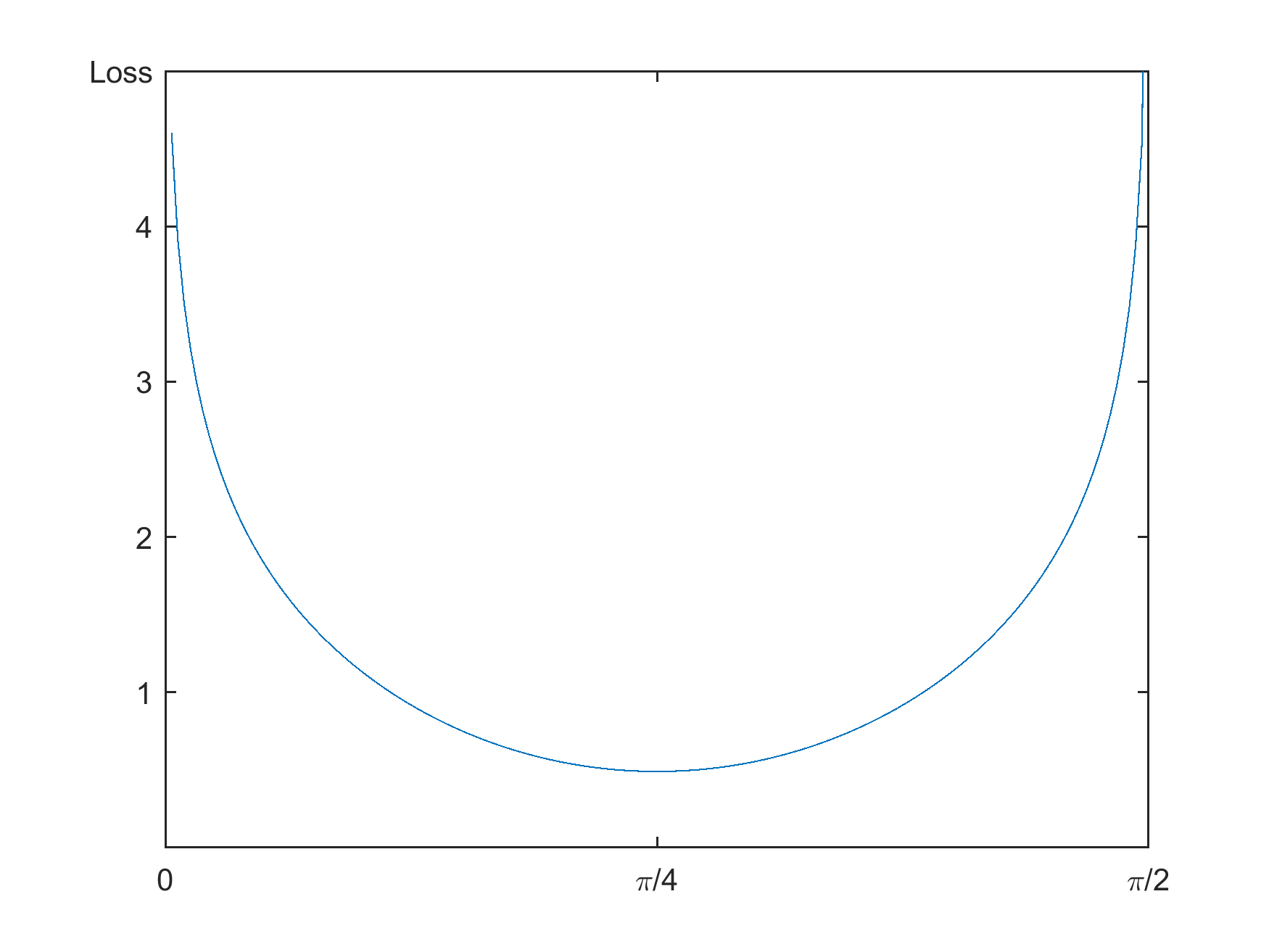}}
\caption{Information loss as defined by Eq.~\ref{informationLossFunction}}
\label{lossFunctionFigure}
\end{center}
\end{figure}

Based on this observation, one can define an information loss function similar to entropy, which measures the expected information for both directions of signal transmission:
\begin{equation}
L_I(d) = \sin(d) \cdot \ln\big(\cos(d)\big) + \cos(d) \cdot \ln\big(\sin(d)\big)
\label{informationLossFunction}
\end{equation}
The information loss function in Eq.~\ref{informationLossFunction} assumes its minimum for an angle of~$45^\circ$, or~$d=\pi/4$, when the signals in both directions are identical, and in equilibrium, as shown in Figure~\ref{lossFunctionFigure}.
Section~\ref{sec:LossFunction} will develop this function into the proposed loss function for training.

\section{Golden ratio}
\label{sec:GoldenRatio}
The golden ratio is a specific ratio between two quantities, which has been known since ancient times, and which has been observed, for example in nature, architecture, and art~\citep{livioGoldenRatioBook}. Informally, two quantities are said to be in the golden ratio if their ratio is the same as the ratio of their sum to the larger of the two quantities. Mathematically, the golden ratio can be derived from the equation developed in the previous section for which the perceived signal is equal to the actual or true signal:
\begin{equation}
p  = \frac{1-p}{p}
\label{goldenRatioEquation}
\end{equation}
A straightforward transformation into the following expression provides one of two common ways of defining the golden ratio as zeros of a quadratic equation:
\begin{equation}
p^2 + p - 1  = 0
\label{plusGoldenRatioEquation}
\end{equation}
This equation has two irrational solutions, namely
\begin{equation}
p_1 = \frac{\sqrt{5}-1}{2} \approx 0.618,
\end{equation}
and
\begin{equation}
p_2 = \frac{-\sqrt{5}-1}{2} \approx -1.618
\end{equation}

Note that the following identity holds for both solutions, $p_1$ and $p_2$, which will become important later in the paper when terms for the learning rate and momentum weight are developed:
\begin{equation}
1-p = p^2
\label{complementPlusEquation}
\end{equation}

The second way of defining the golden ratio is to replace $p$ by $-p$ in Eq.~\ref{plusGoldenRatioEquation}, which leads to this quadratic equation:
\begin{equation}
p^2 - p - 1  = 0
\label{minusGoldenRatioEquation}
\end{equation}
Again, this equation has two solutions, which are the negatively signed solutions of Eq.~\ref{plusGoldenRatioEquation}:  
\begin{equation}
-p_1 \approx -0.618 \mbox{\quad and} -p_2 \approx 1.618
\end{equation}

However, for the second quadratic equation, Eq.~\ref{minusGoldenRatioEquation}, the complement $1-p$ is given by:
\begin{equation}
1-p = -\frac{1}{p}
\label{complementMinusEquation}
\end{equation}
Similar to Eq.~\ref{complementPlusEquation}, this property will be used later in the paper for the development of regularization terms. It turns out that Eq.~\ref{complementPlusEquation} and Eq.~\ref{complementMinusEquation} are closely connected with each other.

Another property of Eq.~\ref{plusGoldenRatioEquation} and Eq.~\ref{minusGoldenRatioEquation} is that the absolute sum of their solutions equals~$1$, respectively:
\begin{equation}
p_1 + p_2 = -1 \mbox{\quad and \quad} -p_1 - p_2 = 1
\label{sumOneEquation}
\end{equation}
In fact, this property is one of the main motivations for this paper. Section~\ref{subsec:Regularization} will argue that exploiting this property can improve machine learning performance. 

Note that most literature refers to the specific number $\varphi \approx 1.618$ as the golden ratio, which was defined by Eq.~\ref{minusGoldenRatioEquation} above. However, this paper will refer to all solutions of Eq.~\ref{plusGoldenRatioEquation} and Eq.~\ref{minusGoldenRatioEquation} as the golden ratio, without singling out a specific number.

\section{Loss function}
\label{sec:LossFunction}
This section will transform the information loss introduced in Section~\ref{sec:ComputationalModel} into a sigmoidal loss function, which will be used for training. Implementing a sigmoidal function is beneficial for a biological system because the output is within fixed limits.

\subsection{Information loss}
\label{subsec:InformationLoss}
Starting from the information loss function given by Eq.~\ref{informationLossFunction} in Section~\ref{sec:ComputationalModel}, let its input~$d$ be computed as follows:
\begin{equation}
d = (y-t+1) \cdot \frac{\pi}{4}
\label{differenceEquation}
\end{equation}
The input~$d$ then denotes the angular difference between the network output~$y$ and the teaching input~$t$, with $d \in \big[0;\frac{\pi}{2}\big]$. This difference is extreme, meaning either $0$~or $\pi/2$, when the difference between the network output and the teaching input is maximum, with $|y-t|=1$. For inputs approaching this maximum difference, the information loss function in Eq.~\ref{informationLossFunction} tends towards infinity. On the other hand, the information loss function attains its minimum when the output and teaching input are identical, which means $d$ is $\pi/4$.  
Eq.~\ref{computationalModelEquation} in Section~\ref{sec:ComputationalModel} computes the expected information based on the true signal~$p$ and the observed signal $(1-p)/p$. Resolving this equation for~$p$ leads to the following sigmoidal function for~$p$:
\begin{equation}
p = \frac{1}{1+\exp\left(-E/p\right)}
\label{sigmoidEquation}
\end{equation}
Inserting the information loss $L_I(d)$ from Eq.~\ref{informationLossFunction} for~$E$ into Eq.~\ref{sigmoidEquation}, and using the equilibrium value~$1/\sqrt{2}$ for~$p$, with $\phi = \pi/4$ in Eq.~\ref{pythagoreanEquation}, produces the following loss function:
\begin{equation}
L(d) = \frac{1}{1+\exp\left(-L_I(d)/\sqrt{2}\right)}
\label{mainLossEquation}
\end{equation}
This loss function reaches its maximum of~$1$ for $d=0$ or $d=\pi/2$, when the distance between the network output and the teaching input is maximum. Its minimum is reached when the output equals the teaching input, for $y-t=0$ and $d=\pi/4$.

 The derivative of the loss function~$L$ with respect to the prediction~$Y$, $dL/dY$, can be computed by applying the chain rule. This leads to the following equation when multiplying the derivative of the outer sigmoid function, which is $L \cdot (1-L)$, and the derivative of $L_I(d)$, and $\partial d / \partial y = \pi/4$:
\begin{equation}
\frac{dL}{dy} = \frac{\pi}{\sqrt{2}^5} \cdot L(d) \cdot \big(1-L(d)\big) \cdot \Bigg(\frac{\sin^2(d)}{\cos(d)} - \frac{\cos^2(d)}{\sin(d)} + \sin(d) \cdot \ln\big(\sin(d)\big) - \cos(d) \cdot \ln\big(\cos(d)\big) \Bigg)
\label{mainLossDerivativeEquation}
\end{equation}
This derivative can then be used for descending the gradient in the traditional backpropagation process. The next section introduces the corresponding learning rate and momentum weight to be used in combination with the loss function in Eq.~\ref{mainLossEquation}) and its derivative in Eq.~\ref{mainLossDerivativeEquation}.

\subsection{Regularization}
\label{subsec:Regularization}
The computational model outlined in Section~\ref{sec:ComputationalModel} implies specific values for the learning rate~$\eta$ and momentum weight~$\alpha$ in Eq.~\ref{deltaWithMomentumEquation} compatible with the model. The reasoning is as follows: The loss function given by Eq.~\ref{mainLossEquation} returns the true signal, which is the gradient of the computational model defined by Eq.~\ref{computationalModelEquation} when regarding the information of the observed signal, $-\ln((1-p)/p)$, as input. Furthermore, according to Eq.~\ref{pythagoreanEquation}, this true signal corresponds to the signal observed in the opposite direction. For the minimum uncertainty, or equilibrium, this signal is equal to $1/\sqrt(2)$. However, for the minimum uncertainty, this signal should be the golden ratio, which is the solution to the model defined by Eq.~\ref{computationalModelEquation}. Therefore, the signal needs to be regularized, using the momentum weight~$\alpha$, so that it satisfies the following requirement:
\begin{equation}
\frac{\alpha}{sqrt(2)} = p_1 \approx 0.618 \quad\Longrightarrow\quad \alpha \approx 0.874
\label{momentumWeightEquation}
\end{equation}
This provides a specific value for the momentum weight, namely~$\alpha \approx 0.874$.

On the other hand, the observed signal also needs to be regulated. This can be accomplished by choosing the learning rate~$\eta$ appropriately, again following a similar reasoning: The observed signal and the true signal, which is the signal observed in the opposite/backwards direction, are in a negatively inverse relation, $p=-1/p$. When viewed from the opposite direction, using Eq.~\ref{complementMinusEquation}, the observed signal in forward direction is equal to one minus the true signal. Therefore, the complement of the signal observed in backward direction needs to be computed in order to obtain the signal in forward direction. According to Eq.~\ref{complementPlusEquation}, this means squaring the signal, which amounts to computing one minus the signal. For this reason, the loss function in Eq.~\ref{mainLossEquation}, which computes the observed signal~$p$ in backward direction based on Eq.~\ref{computationalModelEquation}, needs to be squared, leading to the ultimate loss function proposed in this paper:
\begin{equation}
Loss(d) = \left(\frac{1}{1+\exp\Big(-\big(L_I(d)-min\big)/\sqrt{2}\Big)}\right)^2,
\label{newLossEquation}
\end{equation}
where $min$ is the minimum of the information loss function in Eq.\ref{informationLossFunction}, which the function attains for~$d=\pi/4$. The derivative in Eq.~\ref{mainLossDerivativeEquation} needs to be adjusted accordingly by replacing $L(d)$ with $Loss(d)$ and multiplying by $2 \cdot Loss(d)$, again using the chain rule. Applying the same processing steps to the momentum weight~$\alpha$ then leads to the following expression for the learning rate~$\eta$:
\begin{equation}
\eta = (1-\alpha)^2 \approx 0.016
\label{learningRateEquation}
\end{equation}
This provides the value for the second regularization term, namely the learning rate~$\eta$, with~$\eta \approx 0.016$.

From the above discussion, it follows that the delta learning rule with momentum, as given by Eq.~\ref{deltaWithMomentumEquation}, adds two regularized gradients, each seen from a different direction. This interpretation of the delta rule and momentum differs from the common understanding that applying Eq.~\ref{deltaWithMomentumEquation} smoothens gradient descent by giving more weight to weight changes in previous iterations. Instead, here it is argued that the delta learning rule with momentum considers gradients for two different directions, ensuring that one is not improved at the expense of the other, while descending to an optimum, until the equilibrium is reached. In fact, for the observed reality, gradient descent follows the relationship given in Eq.~\ref{sumOneEquation}, with the sum of both gradients becoming one when the minimum uncertainty is reached in the state of equilibrium. According to the theory laid out here, it is this balancing of gradients that makes the delta rule with momentum so successful in practical applications. In both directions, gradients contribute with the same weight.

\section{Experimental Evaluation}
\label{sec:ExperimentalEvaluation}
\begin{figure}[t] 
\begin{center}
\centerline{\includegraphics[width=0.84\columnwidth]{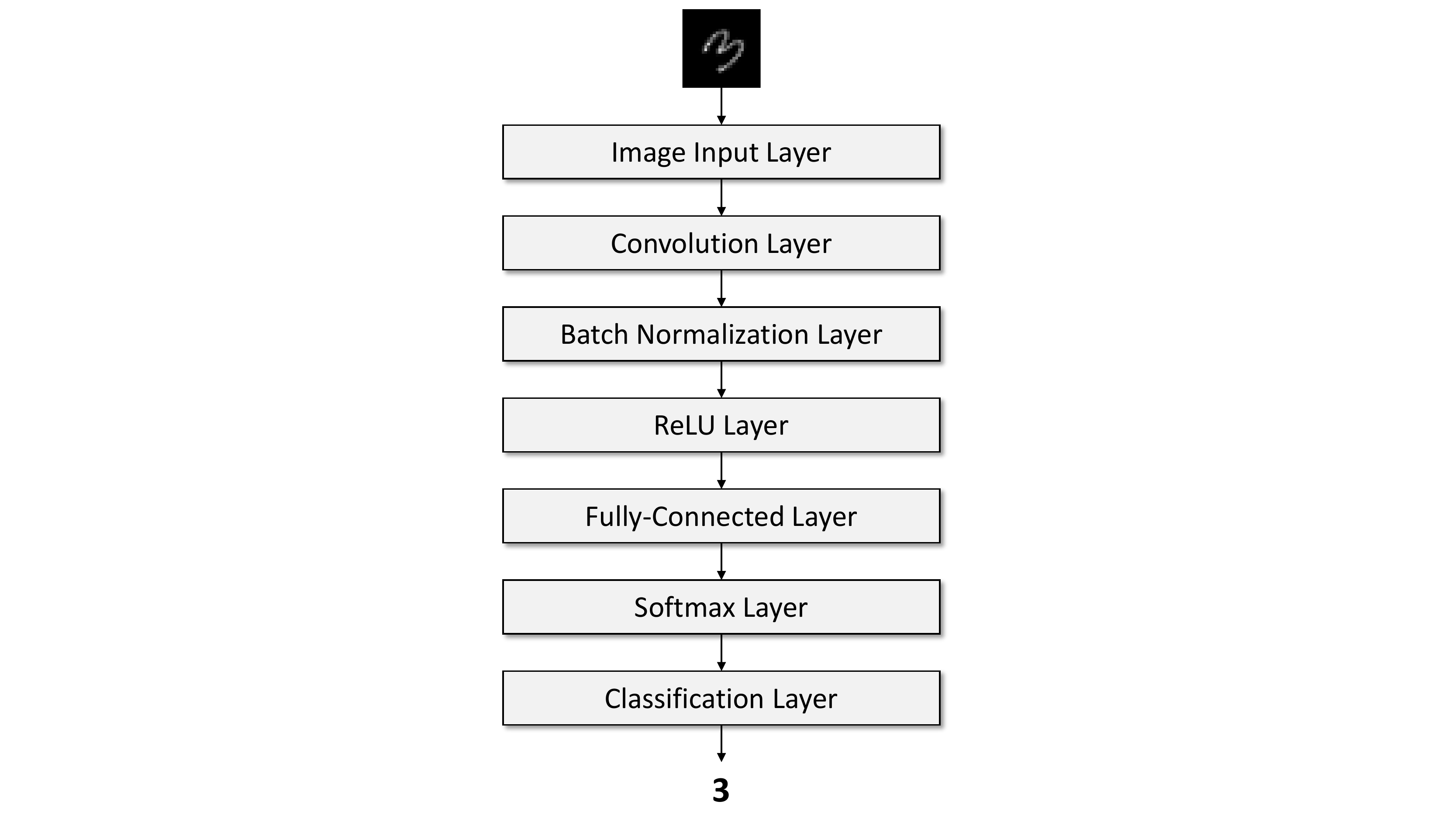}}
\caption{Network architecture}
\label{networkArchitectureFigure}
\end{center}
\end{figure}

To show that the proposed loss function in Eq.~\ref{newLossEquation} works in conjunction with the derived learning rate~$\eta$ and momentum weight~$\alpha$, a practical experiment is performed on public data. For handwritten digit classification, a deep learning network is trained on a dataset containing 10,000 handwritten, artificially rotated digits, and evaluated by averaging ten runs for each fold in 10-fold cross-validation~\citep{DigitDataMATLAB}. Each digit is a 28-by-28 gray-scale image, with a corresponding label denoting which digit the image represents (MNIST database~\citep{MNISTbyYannLecun}). Figure~\ref{networkArchitectureFigure} shows the network architecture used in the experiment, with a sample input digit~$3$ and a correct output result~\citep{krizhevsky2012imagenet}.
The first layer is the image input layer, with a size of 28-by-28, followed by a convolution layer with 20 5-by-5 filters. The next layers are a batch normalization layer, a ReLU layer, and a fully-connected layer with an output size of~$10$. Finally, a softmax layer and a classification layer are the last two layers of the network, with the latter computing the proposed loss function in Eq.~\ref{newLossEquation}. For training, the learning rate given by Eq.~\ref{learningRateEquation} and the momentum weight given by Eq.~\ref{momentumWeightEquation} are used.

Table~\ref{resultsTable} shows the classification results for training with the common loss function defined by the sum of squares, and with the proposed loss function defined by Eq.~\ref{newLossEquation}.
\begin{table}
  \caption{Experimental results with 10-fold cross-validation}
  \label{resultsTable}
  \centering
  \begin{tabular}{ccc}
    \toprule
    Loss & avg. accuracy (\%) & std \\
    \midrule
    SSE without momentum & 77.4 & 10 \\
    SSE with momentum & 98.9 & 1 \\
    Eq.~\ref{newLossEquation} with momentum & 99.4 & 0.1 \\
    \bottomrule
  \end{tabular}
\end{table}
All results have been achieved after~$30$ training epochs, using ten-fold cross-validation. The results show that training with SSE loss benefits significantly from using a momentum term, which increases the accuracy from $77.4\%$ to $98.9\%$. The proposed loss function in Eq.~\ref{newLossEquation} with momentum performs best, with an accuracy of $99.4\%$. It is also worth noting that the standard deviation improves by an order of magnitude each time, decreasing from~$10$ for SSE without momentum to only~$0.01$ for the proposed loss function, learning rate, and momentum weight.

\section{Conclusion}
For training of ANNs, this paper presents an information-theoretical loss function that is motivated by biomedical signal processing via neurotransmitters in a biological synapse. Learning is based on the principle of minimizing uncertainty, as measured by the loss function. Another important aspect is that learning is considered to be a two-directional process between network input and teaching input, with either input becoming the output in one direction. Both inputs become identical in points defined by the golden ratio. Therefore, the golden ratio, which has gained little attention in the machine learning literature so far, takes center stage here. Network weights are adjusted so that the absolute sum of gradients in both directions equals one. This helps in dampening oscillations, while the network is approaching a state of minimum energy. Technically, this is achieved by setting the learning rate and the momentum weight to specific values, thus explaining the generally accepted usefulness of the momentum term in a formal framework. This also confirms empirical values generally used in the literature for learning rate and momentum weight.

To validate this information-theoretical approach further, classification results for a handwritten digit recognition task are presented, showing that the proposed loss function in conjunction with the derived learning rate and momentum weight works in practice.  


\section*{Broader Impact}
The broader impact of this paper lies in the theoretical explanation of the learning rate and momentum term, and in how their values can be determined, in a learning process based on backpropagation. The theoretical findings confirm and specify more precisely the empirical values often used in the literature for practical experiments. This provides a guideline for future experiments, relieving researchers and developers of a tedious parameter search. The paper derives this result by studying the basic neurotransmitter mechanisms in the synaptic cleft between the presynaptic and postsynaptic neuron of a biological synapse. This study also revealed that the golden ratio plays a key role role in neural signal transduction, particularly in synaptic transmission/neurotransmission. Therefore, a full understanding of neural information processing may not be possible without considering the golden ratio.  

\begin{ack}
This work was supported by the Intramural Research Program of the Lister Hill National Center for Biomedical Communications (LHNCBC) at the U.S. National Library of Medicine (NLM), and the National Institutes of Health (NIH).
\end{ack}



\bibliography{NeurIPS_2020-Jaeger-arXiv}
\bibliographystyle{plainnat}

\end{document}